\documentclass{article}
\usepackage[preprint]{colm2026_conference}

\usepackage{amsmath}
\usepackage{booktabs}
\usepackage{graphicx}
\usepackage{microtype}
\usepackage{hyperref}
\usepackage{url}
\usepackage{xcolor}
\usepackage{colortbl}
\usepackage{multirow}

\definecolor{darkblue}{rgb}{0,0,0.5}
\hypersetup{
  colorlinks=true,
  citecolor=darkblue,
  linkcolor=darkblue,
  urlcolor=darkblue,
  pdftitle={Crayotter: Learning Long-Horizon Video Editing Agents via Group-Relative Preference Backpropagation},
  pdfauthor={Lecheng Yan, Jianze Lin, Yichong Zhang, Ben Pan, Wenxi Li, Chenyang Lyu, Liting Zhou, Cathal Gurrin}
}

\title{Crayotter: Learning Long-Horizon Video Editing Agents via Group-Relative Preference Backpropagation}

\author{
\parbox{0.96\textwidth}{\centering
Lecheng Yan$^{1}$ \quad Jianze Lin$^{2}$ \quad Yichong Zhang$^{3}$ \quad Ben Pan$^{4}$\\
Wenxi Li$^{5}$ \quad Chenyang Lyu$^{6}$ \quad Liting Zhou$^{7}$ \quad Cathal Gurrin$^{7}$\\
{\normalfont\small
$^{1}$University of Science and Technology of China \quad $^{2}$Beijing Normal University\\
$^{3}$Jilin University \quad $^{4}$Tianjin University \quad $^{5}$East China Normal University\\
$^{6}$Alibaba Group \quad $^{7}$Dublin City University
}}
}

\begin{document}

\maketitle

\begin{abstract}
Long-horizon video editing agents receive final-product feedback only after many interdependent decisions. Yet editing quality is subjective, admits multiple valid solutions, and is not meaningfully calibrated across heterogeneous requests, making a global scalar objective both ambiguous and temporally uninformative. Our key observation is that fixing the request, materials, and production constraints converts this subjective objective into an ordinal comparison among directly comparable alternatives. We introduce \textbf{Group-Relative Preference Backpropagation (GRPB)}, which transforms same-task rankings into zero-sum advantages and redistributes them as bounded credit over semantic editing segments. A lagged allocator and guarded transmission prevent current judgments or unreliable estimates from directly shaping the same rollout group. We manually construct a project-disjoint, horizon-stratified suite of realistic editing tasks for training and controlled evaluation. Across matched baselines, credit interventions, external benchmarking, and blinded human evaluation, GRPB improves both editing behavior and rendered products. The resulting 9B Crayotter model surpasses several proprietary systems on AgenticVBench, supporting task-local preference reduction as a practical approach to learning from subjective, delayed outcomes. Code and all supporting materials are publicly available at \url{https://github.com/idwts/Crayotter}.
\end{abstract}

\section{Introduction}

An autonomous video editor must inspect source footage, select temporal regions, construct a timeline, render, and revise. Recent systems expose this multistage structure through clip filtering and composition \citep{Yang2024AVT}, editor--critic interaction \citep{SandovalCastaneda2025EditDuet}, and planning over hours-long footage \citep{Zhao2026CutClaw}. They do not, however, resolve how a judgment of the rendered video should train the earlier decisions that produced it.

Execution traces provide verifiable feedback about tool completion, artifact validity, duration, and stage coverage, but not semantic selection, narrative continuity, pacing, or stylistic fidelity. These qualities become observable only after rendering, many decisions after their causes. They are also request dependent: different tasks admit different valid edits and weight quality dimensions differently. An absolute score across tasks therefore entangles product quality, task difficulty, and judge-scale variation.

The key reduction is to condition comparison on the task. Instead of regressing a globally calibrated quality value, we sample alternative trajectories for the same request, materials, target duration, and constraints, and ask only which final edit is preferred. The underlying judgment may remain subjective, but the controlled context makes the candidates directly comparable and yields a concrete ordinal relation. This converts a subjective, multi-solution generation objective into a well-posed within-task preference-learning problem without requiring cross-task score calibration. It also changes the appropriate unit of data construction: training examples should preserve projects and their alternative trajectories as comparison groups, rather than pool independently scored outputs across unrelated tasks. The remaining challenge is temporal: the preference belongs to a final video, whereas policy optimization requires credit at the editing decisions that produced it.

\begin{figure}[t!]
  \centering
  \includegraphics[width=.95\linewidth]{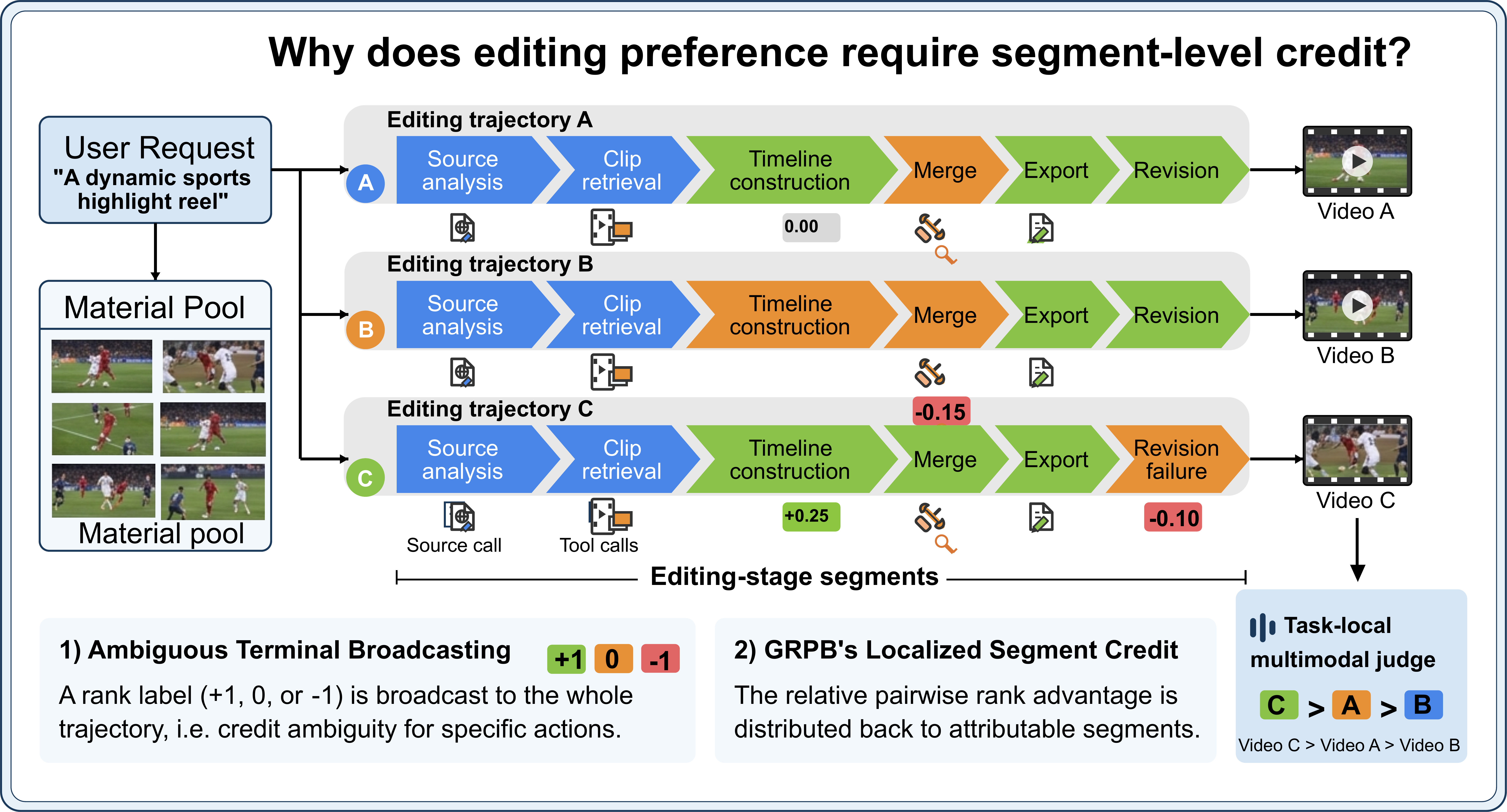}
\caption{Why editing preference needs segment-level credit. Same-task alternative edits make final-product preference meaningful, but terminal rank alone does not identify where policy credit should be placed. GRPB uses a learned, lagged allocator to localize this task-relative signal.}
\label{fig:preference_backtrace}
\end{figure}

Return redistribution addresses delayed outcomes \citep{arjona2019rudder}, agentic methods estimate progress toward verifiable goals \citep{spaRL25,agentPRM25}, and preference-based RL replaces scalar specification with behavioral comparisons \citep{christiano2017deep}. None directly resolves the combination needed here: task-local subjective outcomes whose credit must be assigned to structured editing stages.

We introduce \textbf{Group-Relative Preference Backpropagation (GRPB)}. For each request, GRPB orders rollouts produced from the same materials and constraints, converts the order into a zero-sum rank advantage, and allocates it over contiguous editing segments with a sparse Bradley--Terry model. Policy credit uses the allocator before it learns from the current group. Reliability gating, per-segment caps, and exact return conservation bound early estimates, while raw judge magnitudes never enter the policy objective.

We instantiate GRPB in the Crayotter editing environment \citep{yan2026crayotter}. To support training and controlled evaluation, we manually collect source materials from realistic editing projects and construct a horizon-stratified task suite with project-disjoint splits. We evaluate the resulting 9B policy through matched training comparisons, an external post-production benchmark, controlled segment-feature interventions, and blinded human preferences.

\textbf{Our contributions are three-fold.}
\begin{enumerate}
\item We formulate long-horizon video editing as \emph{task-local artifact preference credit assignment}, reducing uncalibrated cross-task quality scores to directly comparable within-task preferences while separating verifiable process feedback from subjective final-product assessment.
\item We propose GRPB, which combines zero-sum within-task rank advantages with a lagged Bradley--Terry segment allocator, reliability gating, capped allocation, and exact return conservation.
\item We manually construct a project-disjoint suite of realistic editing tasks across three horizon levels. Matched 9B experiments demonstrate gains in post-production performance, credit localization, and blinded human preference.
\end{enumerate}

\section{Related Work}

\subsection{Relative Outcomes and Process Credit}

Delayed credit assignment is a longstanding reinforcement-learning problem. Reward shaping introduces intermediate feedback under policy-invariance conditions \citep{ng1999policy}, whereas RUDDER redistributes delayed returns toward influential events \citep{arjona2019rudder}. PPO and GAE stabilize optimization once rewards have been defined \citep{schulman2017ppo,schulman2016gae}, but do not determine how a judgment of a completed artifact should be assigned to the stages that produced it.

Recent agent-learning methods decompose interaction traces or infer stepwise progress from terminal supervision \citep{luo2025agentlightning,spaRL25,hcapo26,agentPRM25}. Related outcome-to-process methods derive process rewards without step annotations \citep{freePRM24,prime25,spPRM25}. Most assume an externally verifiable terminal condition. Group-relative optimization instead avoids a global reward scale by comparing responses to the same prompt \citep{deepseekMath25}, but ordinarily broadcasts one trajectory-level advantage to every action. GRPB combines these ideas differently: it first constructs a zero-sum ordinal advantage within a task group, then learns how to conserve and redistribute that advantage over semantically distinct segments.

\subsection{Preference Learning for Subjective Artifacts}

Preference-based RL replaces a difficult scalar objective with comparisons of behaviors or ranked trajectories \citep{christiano2017deep,brown2019trex,hindsightPrior24}. This is especially relevant when valid outputs are diverse and quality is not comparable across tasks. GRPB restricts each comparison to rollouts sharing the same request, materials, and constraints, thereby reducing subjective final-product assessment to a task-local ordering problem. Unlike methods that treat an outcome label as dense step correctness, it infers relative segment attribution. Because learned proxies can decouple from held-out quality and model judges exhibit systematic biases \citep{gao2023scaling,zheng2023judging,llmJudge25}, GRPB discards raw judge magnitudes, uses a pre-update allocator for policy credit, and bounds the amount assigned to any segment.

Video editing provides a concrete instance of this setting: existing systems formulate trimming, nonlinear editing, and long-footage composition as multistage agent workflows \citep{Yang2024AVT,SandovalCastaneda2025EditDuet,Zhao2026CutClaw,li2026direct,zhang2026benchmark}, while prior RL work optimizes sequential editing decisions \citep{rlVideoEditing24}. Artifact evaluators can estimate dimensions of rendered-video quality but remain imperfect proxies \citep{videoScore24,evalCrafter24}. GRPB addresses the complementary learning problem of connecting task-local judgments of rendered artifacts to reliable intermediate policy credit.

\begin{figure}[t]
  \centering
  \includegraphics[width=.9\textwidth]{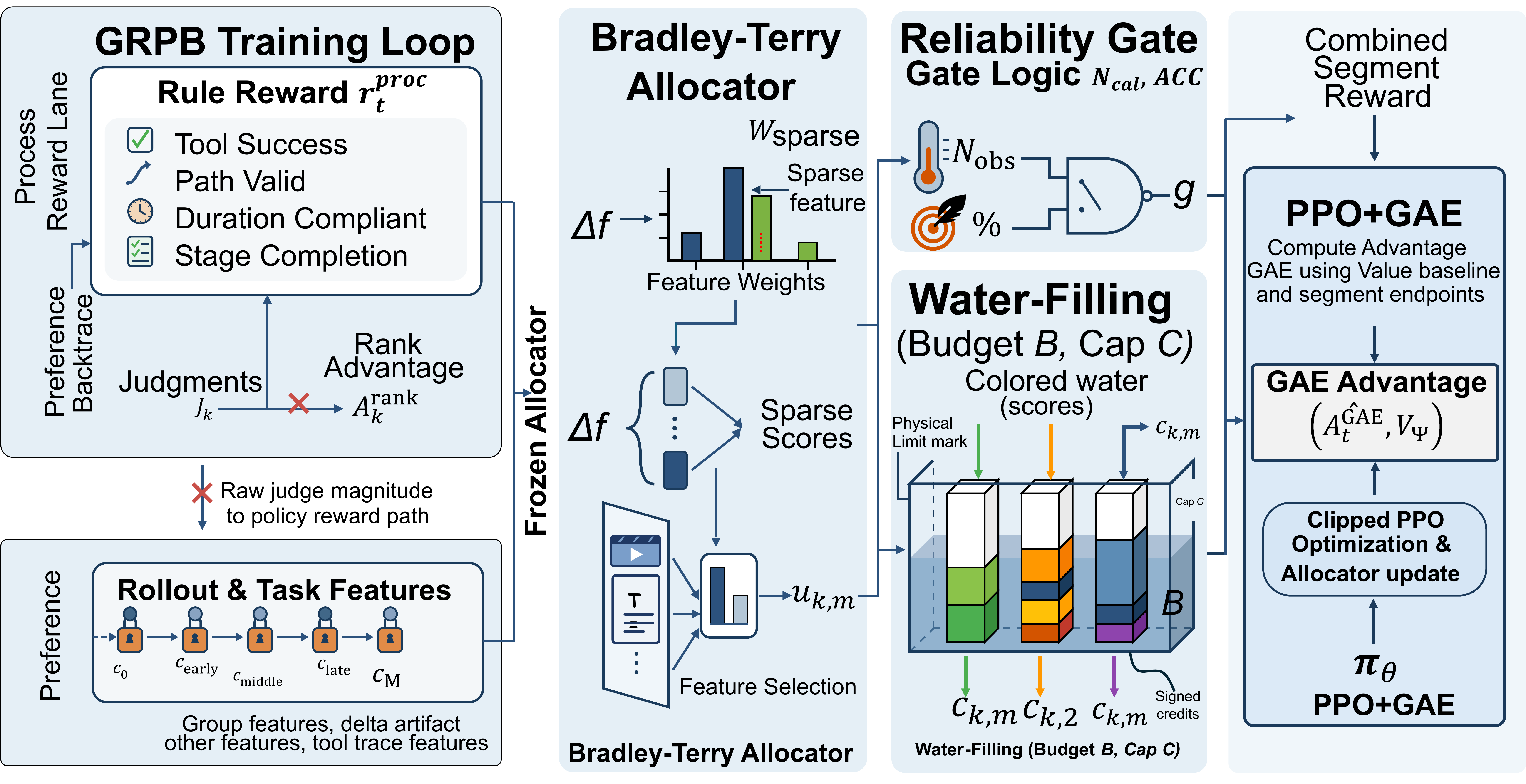}
\caption{GRPB training architecture. Executable process feedback and task-relative final-product preference meet at segment endpoints. Policy credit uses the pre-update allocator; same-group comparisons train only the allocator used by subsequent batches.}
  \label{fig:backtrace_training_loop}
\end{figure}

\section{Method: Group-Relative Preference Backpropagation}

We instantiate GRPB in the Crayotter editing environment \citep{yan2026crayotter}, while formulating the method over an artifact-observable interface rather than a system-specific workflow.

\subsection{Crayotter Editing Environment}

The Crayotter editing environment presents the agent with a request, source-material pool, multimodal analyses, editing blueprint, and workspace artifacts. Structured analysis, timeline, inspection, repair, and export calls update the workspace while recording execution and artifact diagnostics. GRPB operates on four observables---segmented traces, rendered outcomes, process diagnostics, and same-task rollout groups---provided by this environment.

\subsection{Problem Formulation and Process Return}

We model editing as a finite-horizon augmented MDP
\[
\mathcal{M}=(\mathcal{S},\mathcal{A},P,\rho_0,r,\gamma,T).
\]
The state $s_t$ summarizes the request and constraints, artifact tree, blueprint coverage, workspace, tool history, rendered previews, and prior video for revision tasks; $a_t$ is a structured editing-tool call.

For a fixed task $q$, we sample a rollout group $\mathcal{G}_q=\{\tau^{(k)}\}_{k=1}^{K}$ whose members share the same request, fixture, materials, target duration, and constraints. Only a valid export is preference-eligible. A product-only multimodal judge observes the request and ordered frames of the final video and returns $J_k\in[0,100]$; it does not observe the hidden tool trace or blueprint. $J_k$ is used only for within-group order, never as a calibrated reward magnitude.

The process channel
\[
r_t^{\mathrm{proc}}=\rho_{\mathrm{proc}}(s_t,a_t,s_{t+1}),
\]
scores tool and artifact validity, export completion, duration, and stage progress. It supports executable rollouts, while final-product preference covers qualities observable only after rendering. Consecutive events with the same editing-stage label form a segment; process residuals and preference credit are placed at segment endpoints.

\paragraph{Artifact-level interface.} Process diagnostics verify executable progress but cannot judge narrative rhythm, continuity, or stylistic fidelity; final-product preference captures these qualities but does not identify their causes. Persistent workspace artifacts bridge the two: contiguous events that transform the same editing stage define a segment to which delayed preference can be returned. GRPB therefore requires observable intermediate artifacts, meaningful segmentation, and same-task alternative trajectories rather than the Crayotter agent's internal organization.

\subsection{Group-Relative Rank Advantage}

For an eligible group of at least $K_{\min}$ exports, GRPB converts judge scores to pairwise win/loss advantage with tie tolerance $\epsilon_J$:
\[
A_k^{\mathrm{rank}}=\frac{1}{K-1}\sum_{j\ne k}
\left[\mathbf{1}(J_k-J_j>\epsilon_J)-
\mathbf{1}(J_j-J_k>\epsilon_J)\right].
\]
This statistic depends only on within-task order, lies in $[-1,1]$, and satisfies $\sum_k A_k^{\mathrm{rank}}=0$ exactly. Thus score offsets, scale changes that do not alter ties, and cross-task calibration do not affect the policy signal. When rollouts share an explicitly recorded counterfactual prefix, only segments after the branch point are attributable; otherwise the full trajectory is used.

\subsection{Lagged Pairwise Segment Allocator}

For segment $m$ of rollout $k$, a sparse vector $f_{k,m}$ encodes stage identity, execution and artifact statistics, trajectory position, repair evidence, request alignment, and optional semantic deltas. We define its predictive contribution $u_{k,m}=w^\top f_{k,m}$ and aggregate rollout utility $\hat z_k=M_k^{-1}\sum_m u_{k,m}=w^\top\bar f_k$, where $\bar f_k=M_k^{-1}\sum_m f_{k,m}$. A Bradley--Terry model \citep{bradley1952rank} predicts
\[
p_w(k\succ j)=\sigma(\hat z_k-\hat z_j).
\]
For every non-tied ordered pair in the current group, the allocator minimizes
\[
\mathcal{L}_{\mathrm{BT}}(w)=
-\sum_{J_k-J_j>\epsilon_J}\log p_w(k\succ j)
+\frac{\lambda_w}{2}\lVert w\rVert_2^2,
\]
using online AdaGrad with clipped weights. This is weak supervision at the rollout level: it assumes that predicted final-product utility can be represented by the mean of segment feature contributions. Under this additive model, the regularized objective selects a unique $w$ and $u_{k,m}$ is exactly the term contributed by segment $m$ to $\hat z_k$, so the score used for allocation is algebraically consistent with the score used for ranking. We interpret $u_{k,m}$ as predictive policy credit rather than an independently observed causal effect; Section~\ref{sec:credit_analysis} evaluates whether this learned decomposition localizes held-out single-segment feature interventions.

The update order is essential. Let $w^-$ be the state loaded before observing the current group. GRPB first computes all policy credit with $w^-$, then evaluates its pre-update ranking accuracy, and only afterward updates and saves $w$. Consequently, current labels cannot be memorized and immediately returned as credit to the same batch. Validation freezes both allocator updates and policy-side preference credit.

\subsection{Reliability-Gated Preference Budget}

Early allocator estimates are gated by a reliability coefficient. With $N_{\mathrm{cal}}$ accumulated pre-update pair observations, warm-up target $N_{\mathrm{warm}}$, exponentially averaged accuracy $a$, and target accuracy $a_*$, we use
\[
g=\min\!\left(1,\sqrt{\frac{N_{\mathrm{cal}}}{8N_{\mathrm{warm}}}}\right)
\operatorname{clip}\!\left(\frac{a-0.5}{a_*-0.5},0,1\right).
\]
Credit is disabled before warm-up or when $g$ is below a threshold. Otherwise, the shared group budget is
\[
B=g\min(B_{\max},M_{\min}C),\qquad b_k=A_k^{\mathrm{rank}}B,
\]
where $M_{\min}$ is the minimum attributable segment count and $C$ is the per-segment absolute cap. Scaling by the least segment-rich rollout ensures that every $b_k$ is feasible without giving longer traces a larger budget.

\begin{figure}[t]
  \centering
  \includegraphics[width=\linewidth]{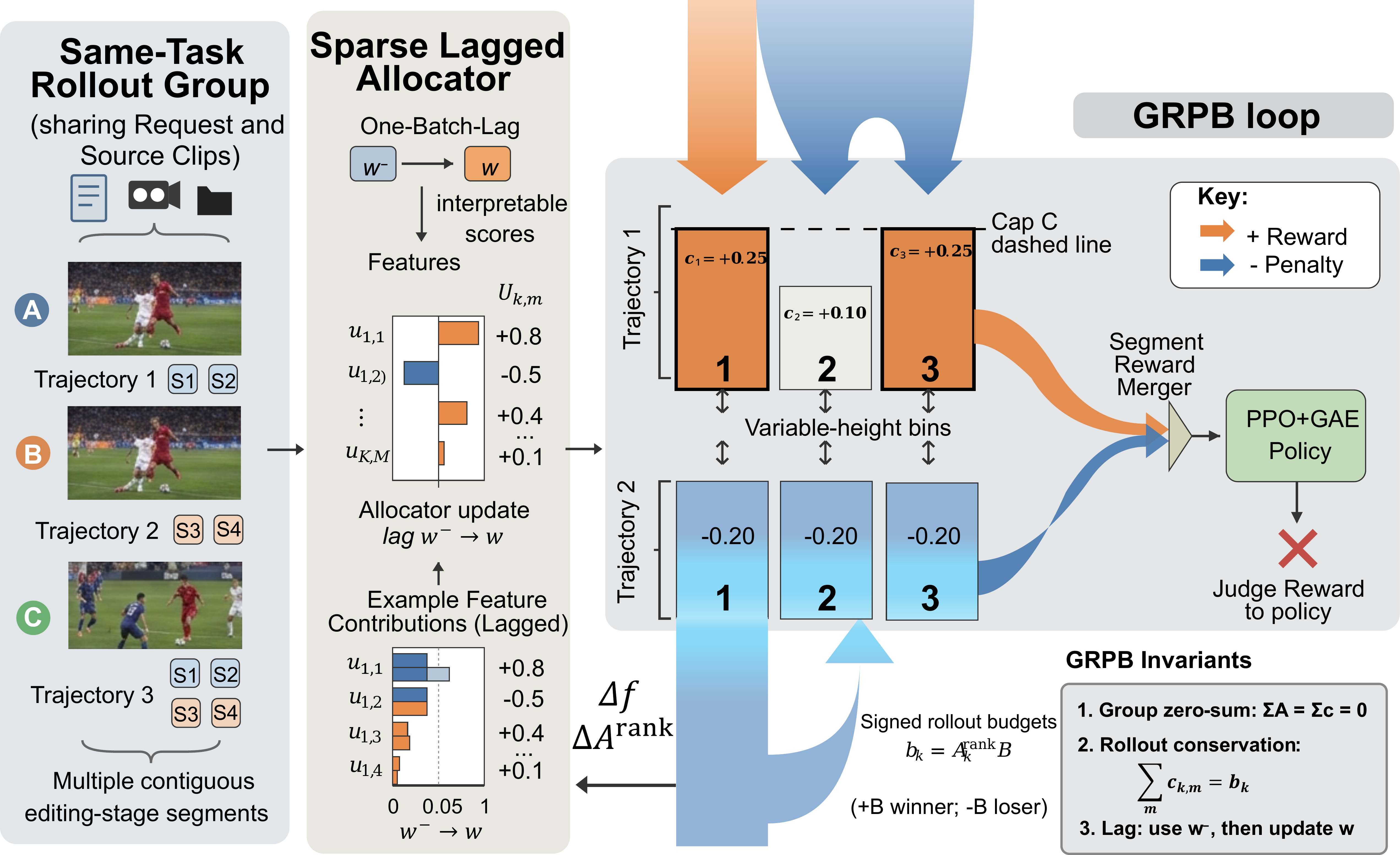}
\caption{Lagged and bounded segment allocation. The sparse pre-update Bradley--Terry model produces feature-additive segment scores. Signed capped allocation conserves each rollout budget and yields exact zero-sum preference pressure across the task-local group.}
  \label{fig:allocator_ig}
\end{figure}

\subsection{Signed Capped Segment Allocation}

Let $s_k=\operatorname{sgn}(b_k)$. For each segment, define
\[
q_{k,m}=\exp\!\left(s_k u_{k,m}/\tau_{\mathrm{alloc}}\right),
\qquad u_{k,m}=(w^-)^\top f_{k,m},
\]
using the allocator frozen before the current rollout group. For $b_k\ne0$ and nonconstant scores, GRPB applies
\[
c_{k,m}=s_k\min\{C,\lambda_kq_{k,m}\},\qquad
\sum_m\min\{C,\lambda_kq_{k,m}\}=|b_k|,
\]
where $\lambda_k$ is the smallest nonnegative solution. The implementation finds it by water-filling: fix every segment whose proportional share reaches $C$, remove it from the active set, and redistribute the remaining mass according to the unchanged $q_{k,m}$. Thus positive budgets favor high scores, whereas negative budgets assign larger-magnitude penalties to low scores. Tied segments receive equal shares; if all scores in any rollout are equal within $10^{-8}$, group credit is withheld rather than uniformly broadcast.

\paragraph{Conservation.} Feasibility follows from $|b_k|\le B\le M_{\min}C\le M_kC$. The function $\sum_m\min\{C,\lambda q_{k,m}\}$ is continuous and nondecreasing from $0$ to $M_kC$, so a solution exists and obeys $|c_{k,m}|\le C$ and $\sum_m c_{k,m}=b_k$. Group-level zero-sum then follows directly:
\[
\sum_k\sum_m c_{k,m}=\sum_k b_k
=B\sum_k A_k^{\mathrm{rank}}=0.
\]

\paragraph{Reward placement.} Credits are stored to six decimals, with any rounding residual assigned to an unsaturated segment so that conservation remains exact. Allocation is over semantic editing segments rather than token counts: $c_{k,m}$ is placed at the final trainable assistant position associated with that segment, preventing long verbal spans from receiving more preference mass merely because they contain more tokens. The rule-based return recorded for the trajectory is unchanged, while the policy-side return gains $b_k$. Raw judge scores remain available for audit but never enter the reward tensor.

\subsection{Critic-Based Policy Optimization}

If segment $m$ of rollout $k$ ends at step $t$, its credit is added to the process reward:
\[
r_t^{(k)}=r_t^{\mathrm{proc},(k)}
  +\sum_{m:\,e(k,m)=t}c_{k,m}.
\]
Given a value function $V_\psi$, we compute
\[
\delta_t=r_t+\gamma V_\psi(s_{t+1})-V_\psi(s_t),
\quad
\hat{A}_t^{\mathrm{GAE}}=
\sum_{\ell=0}^{T-1-t}(\gamma\lambda_{\mathrm{GAE}})^\ell\delta_{t+\ell}.
\]
The policy is updated with the clipped PPO surrogate \citep{schulman2017ppo}
\[
\mathcal{L}_{\mathrm{CLIP}}(\theta)=
\mathbb{E}_t\!\left[
\min\!\left(
\rho_t(\theta)\hat{A}_t^{\mathrm{GAE}},
\operatorname{clip}\!\left(\rho_t(\theta),1-\epsilon,1+\epsilon\right)
\hat{A}_t^{\mathrm{GAE}}
\right)\right],
\]
where $\rho_t(\theta)=\pi_\theta(a_t\mid s_t)/\pi_{\theta_{\mathrm{old}}}(a_t\mid s_t)$. The distinction from GRPO is not the clipped likelihood ratio, which both methods use. GRPB retains the learned value baseline and applies GAE \citep{schulman2016gae} to obtain step-varying advantages from the localized segment rewards; the rollout group is used only to construct preference budgets. GRPO instead replaces the critic-based advantage with a trajectory-level group-relative scalar that is ordinarily shared by all actions in the response. Complete implementation and optimization configurations are provided in the Supplementary Material.

\section{Experiments}

\subsection{Experimental Setup}
\label{sec:experimental_setup}

\paragraph{Training and evaluation data.} We manually collect source materials for 23 realistic post-production projects and construct each project with a user brief, target duration, multimodal analyses, and an executable workspace; revision tasks additionally include an earlier render and natural-language feedback. The resulting 70 tasks comprise 12 normal, 12 medium-horizon, and 46 long-horizon cases, progressing from single-pass production to multi-constraint composition and iterative revision through increased stage coverage, dependency, and revision depth. We split projects before task construction: 18 yield 54 training tasks (9/9/36 by horizon) and five yield 16 held-out tasks (3/3/10), with no request, footage, render, or revision chain crossing partitions. Alternative rollouts within a task share the complete production context and form the comparison groups required by GRPB. AgenticVBench \citep{cao2026agenticvbench} is a separate post-training benchmark.

\paragraph{Matched protocol.} All variants share the same 9B initialization, tool interface, rollout conditions, process supervision, product judge, and optimization budget. Process feedback rewards executable progress, whereas the compared methods differ only in how same-task final-product preference is converted into policy credit. Evaluation combines matched baselines and ablations, the disjoint external benchmark, controlled credit-localization interventions, and blinded human preference.

\begin{figure}[t]
  \centering
  \includegraphics[width=\textwidth]{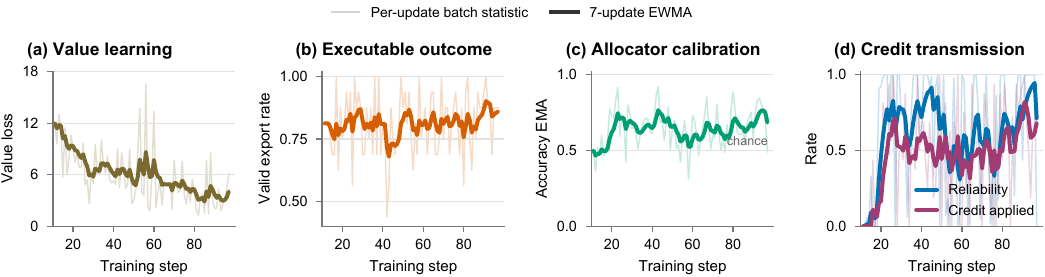}
  \caption{GRPB training dynamics. Thin curves are per-update statistics; thick curves are trailing seven-update EWMAs. Panels show (a) value loss, (b) valid export rate, (c) frozen-allocator accuracy before the update, and (d) reliability versus applied preference credit.}
  \label{fig:grpb_training_dynamics}
\end{figure}

\begin{table}[t]
\centering
\small
\textbf{(a) AgenticVBench results (\%)}\par\smallskip
\begin{tabular*}{\textwidth}{@{\extracolsep{\fill}}lc>{\columncolor{black!7}}cccc@{}}
\toprule
Method & Average & Repurpose & Sequencing & Repair & Assembly \\
\midrule
Base & 1.0 & 1.5 & 0.8 & 1.2 & 0.0 \\
Process-PPO & 13.7 & 20.2 & 7.3 & 15.2 & 9.0 \\
Terminal Rank PPO & 12.0 & 16.5 & 7.4 & 15.4 & 6.9 \\
\addlinespace[2pt]
\multicolumn{6}{l}{\textit{GRPB variants}} \\
\quad uniform allocation & 11.0 & 16.2 & 7.7 & 9.0 & 7.5 \\
\quad without one-group lag & 12.5 & 19.1 & 7.8 & 9.6 & 9.6 \\
\quad without reliability/cap safeguards & 10.9 & 16.6 & 7.7 & 8.9 & 6.4 \\
GRPB (ours) & 15.7 & 23.0 & 7.9 & 15.2 & 14.0 \\
\bottomrule
\end{tabular*}

\medskip
\textbf{(b) Training means over Steps 11--100 (\%)}\par\smallskip
\begin{tabular*}{\textwidth}{@{\extracolsep{\fill}}lccc@{}}
\toprule
Method & Pre-update accuracy & Credit coverage & Export success \\
\midrule
Base & -- & -- & -- \\
Process-PPO & -- & -- & 6.7 \\
Terminal Rank PPO & 50.0 & 11.0 & 21.3 \\
\addlinespace[2pt]
\multicolumn{4}{l}{\textit{GRPB variants}} \\
\quad uniform allocation & 69.3 & 0.3 & 6.9 \\
\quad without one-group lag & 46.6 & 0.0 & 1.7 \\
\quad without reliability/cap safeguards & 63.4 & 0.3 & 17.4 \\
GRPB (ours) & 64.5 & 49.4 & 81.3 \\
\bottomrule
\end{tabular*}
\caption{Matched 9B comparison. AgenticVBench values are final-checkpoint scores; training statistics are averaged over Steps~11--100. Credit coverage is the fraction of episodes receiving nonzero preference credit.}
\label{tab:controlled_main}
\end{table}

\subsection{Training Dynamics}

Figure~\ref{fig:grpb_training_dynamics} separates critic fitting, executable output, allocator calibration, and credit transmission. Value loss declines while valid export remains high across varying task groups. Pre-update allocator accuracy stays above chance over most updates, excluding same-batch fitting as its explanation. Reliability and applied credit rise as calibration evidence accumulates; their gap reflects the additional validity, ranking, and segment-contrast conditions required before preference reaches the policy.

\subsection{Comparative Evaluation and Ablations}

Base denotes the original 9B weights. Process-PPO uses process reward only; Terminal Rank PPO places rank credit at the trajectory endpoint; and uniform allocation divides it over segments. The no-lag and no-safeguard variants respectively remove pre-update allocation and guarded transmission. All trained variants otherwise use matched conditions.

GRPB obtains the highest matched AVG (15.7) and Repurpose score (23.0), exceeding Process-PPO by 2.0 and 2.8 points. Terminal and uniform credit both underperform, showing that task-local preference must be localized. Removing lagging or safeguards also falls below Process-PPO, whereas full GRPB credits 49.4\% of episodes while retaining 81.3\% valid export. The gain therefore requires differentiated and guarded credit.

\paragraph{External benchmark.} AgenticVBench~\citep{cao2026agenticvbench} evaluates repurposing, sequencing, repair, and assembly on 100 disjoint tasks. Table~\ref{tab:agentic_vbench} follows its official rubrics, weighting, and judges. Because model and harness both vary, it measures external performance rather than isolating GRPB.

\begin{table}[t]
\centering
\small
\newcommand{\agentharnesscell}[2]{%
  \begin{tabular}[c]{@{}l@{}}%
    \vphantom{\textbf{Crayotter-9B}}#1\\%
    \hspace{0.8em}\vphantom{\textit{\textbf{Crayotter env.}}}\textit{#2}%
  \end{tabular}}
\textbf{(a) Closed-source models}\par\smallskip
\begin{tabular*}{\textwidth}{@{\extracolsep{\fill}}lc>{\columncolor{black!7}}cccc@{}}
\toprule
Agent / Harness & AVG & \textbf{Rep.} & Seq. & Repair & Asm. \\
\midrule
\agentharnesscell{GPT-5.6-sol}{Codex} & 38.4 & 34.0 & 33.0 & 39.0 & 48.0 \\
\agentharnesscell{Gemini 3.1 Pro}{OpenCode} & 23.8 & 23.0 & 19.0 & 20.0 & 33.0 \\
\agentharnesscell{GPT-5.4-mini}{Codex} & 14.5 & 24.0 & 6.0 & 13.0 & 16.0 \\
\agentharnesscell{Claude Sonnet 4.6}{OpenCode} & 14.4 & 26.0 & 9.0 & 6.0 & 17.0 \\
\agentharnesscell{Gemini 3 Flash}{Gemini CLI} & 13.6 & 20.0 & 3.0 & 12.0 & 19.0 \\
\agentharnesscell{GPT-5.4-mini}{OpenClaw} & 7.7 & 14.0 & 3.0 & 7.0 & 7.0 \\
\bottomrule
\end{tabular*}

\medskip
\textbf{(b) Open-source models}\par\smallskip
\begin{tabular*}{\textwidth}{@{\extracolsep{\fill}}lc>{\columncolor{black!7}}cccc@{}}
\toprule
Agent / Harness & AVG & \textbf{Rep.} & Seq. & Repair & Asm. \\
\midrule
\agentharnesscell{\textbf{Crayotter-9B}}{\textbf{Crayotter env.}}
  & \textbf{15.7} & \textbf{23.0} & \textbf{7.9} & \textbf{15.2} & \textbf{14.0} \\
\agentharnesscell{Qwen3.5-397B-A17B}{Crayotter env.} & 10.9 & 15.3 & 8.1 & 9.6 & 7.5 \\
\agentharnesscell{Kimi-K3-2.8T}{Crayotter env.} & 9.3 & 15.7 & 5.8 & 8.2 & 7.5 \\
\agentharnesscell{Gemma-4-31B-it}{Crayotter env.} & 7.9 & 14.5 & 4.5 & 8.9 & 3.6 \\
\agentharnesscell{Qwen3-VL-235B}{OpenCode} & 3.0 & 4.0 & 1.0 & 6.0 & 1.0 \\
\agentharnesscell{Qwen3.5-9B}{Crayotter env.} & 1.0 & 1.5 & 0.8 & 1.2 & 0.0 \\
\bottomrule
\end{tabular*}
\caption{AgenticVBench results (\%) for representative closed- and open-source models. Crayotter-9B is our 9B model trained with GRPB. All scores follow the official Pillar~0--4 rubrics, judge models (Claude Opus~4.7 and Gemini~3.1~Pro), and 100-task weighting. Repurpose, the most editing-aligned pillar, is shaded.}
\label{tab:agentic_vbench}
\end{table}

Crayotter-9B ranks third among the listed systems~\citep{gemmateam2026gemma4,qwen3technicalreport,qwen3.5,kimiteam2026kimik3openfrontier,bai2025qwen3} and exceeds several proprietary alternatives. It matches Gemini 3.1 Pro on Repurpose but remains weaker on sequencing. Together, Table~\ref{tab:controlled_main} attributes the gain to localized, guarded credit, while Table~\ref{tab:agentic_vbench} shows transfer beyond the training environment without claiming uniform improvement across post-production skills.

\subsection{Credit-Assignment Analysis}
\label{sec:credit_analysis}

We construct 168 evaluation-only controlled pairs from 64 trajectories in the five held-out projects. Each pair degrades only one segment's execution, artifact, and semantic features while fixing the request and all other segments; the probes never update the allocator. They cover material selection, temporal assembly, narration/subtitle alignment, and final checks. Top-1 localization and target mass measure where credit is placed, while pairwise accuracy and log loss test whether the frozen predictor orders the favorable record above its degraded counterpart. The diagnostic bypasses reliability gating to evaluate the learned score but retains the common cap and conservation projection.

GRPB raises Top-1 localization from 11.9\% to 20.8\% and reaches 85.7\% pairwise accuracy without cap violations. Uniform allocation cannot distinguish the intervened segment, while terminal broadcast places reward at an unrelated endpoint and violates the cap by construction. GRPB improves both localization and ordering while preserving the exact budget. The smaller changes in target mass and log loss indicate that identifying a useful credit direction is easier than calibrating its full distribution, motivating reliability gating and bounded allocation on the policy path.

\subsection{Blinded Human Preference Evaluation}

Three human judges independently compare anonymized, randomly ordered outputs from identical held-out requests, sources, and decoding budgets, considering instruction satisfaction, content selection, coherence, fluency, and audiovisual quality. Win and loss rates are averaged per judge to weight evaluators equally, and net win rate is reported from the GRPB perspective.

\begin{table}[t]
\centering
\small
\setlength{\tabcolsep}{2.0pt}
\renewcommand{\arraystretch}{1.08}
\begin{tabular*}{\columnwidth}{@{\extracolsep{\fill}}lccccc@{}}
\toprule
\multirow{2}{*}{Method} &
  Top-1 & Target & Pairwise & Log & Cap \\
&
  loc. (\%) & mass (\%) & acc. (\%) & loss & viol. (\%) \\
\midrule
Terminal broadcast & 1.8 & 1.8 & 50.0 & 0.693 & 100.0 \\
Uniform allocation & 11.9 & 11.9 & 50.0 & 0.693 & 0.0 \\
GRPB (ours) & \textbf{20.8} & \textbf{12.3} & \textbf{85.7} & \textbf{0.670} & 0.0 \\
\bottomrule
\end{tabular*}
\caption{Credit localization on 168 held-out single-segment feature interventions. Target mass is the fraction of absolute credit assigned to the changed segment. Conservation error is zero for all methods.}
\label{tab:credit_localization}
\end{table}

\begin{table}[t]
\centering
\small
\setlength{\tabcolsep}{3.5pt}
\renewcommand{\arraystretch}{1.08}
\begin{tabular*}{\columnwidth}{@{\extracolsep{\fill}}lccc@{}}
\toprule
Opponent & Win (\%) & Loss (\%) & Net win rate $\uparrow$ (\%) \\
\midrule
Base & 67.1 & 32.9 & 34.2  \\
Process-PPO & 54.0 & 46.0 & 8.0 \\
Terminal Rank PPO & 51.6 & 48.4 & 3.2 \\
Uniform Allocation & 55.9 & 44.1 & 11.8 \\
\bottomrule
\end{tabular*}
\caption{Blinded same-task preference from the GRPB perspective, averaged across three human judges.}
\label{tab:artifact_preference}
\end{table}

GRPB achieves positive net win rates against every opponent: 34.2 points over Base, 8.0 over Process-PPO, 3.2 over Terminal Rank PPO, and 11.8 over uniform allocation. The gain over the original weights confirms improved rendered products, while the margins over Terminal Rank PPO and uniform allocation support task-local ordering and differentiated segment credit.

\section{Conclusion}

We introduced GRPB, a preference-to-process reinforcement learning method that converts same-task comparisons into zero-sum rank advantages, localizes them through a lagged and bounded segment allocator, and optimizes the resulting credit with critic-based PPO and GAE. Matched experiments, controlled credit interventions, external benchmarking, and blinded human evaluation show that localized preference credit improves agent behavior and rendered-video quality; Crayotter-9B also ranks above several proprietary systems on AgenticVBench. More broadly, task-local comparison, conservative credit allocation, and artifact-level evaluation provide a practical foundation for other subjective, multi-solution, multistage production tasks.

\bibliographystyle{colm2026_conference}
\bibliography{references}

\end{document}